%% file: conference_101719.tex
\documentclass[conference]{IEEEtran}
\IEEEoverridecommandlockouts
\usepackage{cite}
\usepackage{amsmath,amssymb,amsfonts}
\usepackage{algorithmic}
\usepackage{graphicx}
\usepackage{textcomp}
\usepackage[utf8]{inputenc}
\usepackage[T1]{fontenc}
\usepackage[english]{babel}
\usepackage{float} 
\usepackage{multirow}
\usepackage[table,xcdraw]{xcolor}
\usepackage{colortbl}
\usepackage{hhline}
\usepackage{enumitem}

\def\BibTeX{{\rm B\kern-.05em{\sc i\kern-.025em b}\kern-.08em
    T\kern-.1667em\lower.7ex\hbox{E}\kern-.125emX}}
\begin{document}
\title{MolFusion: Multimodal Fusion Learning for Molecular Representations via Multi-granularity Views
}

\author{\IEEEauthorblockN{1\textsuperscript{st} Muzhen Cai}
\IEEEauthorblockA{\textit{Faculty of Computing,} \\
\textit{Harbin Institute of Technology}\\
Harbin, China \\
mzcai@ir.hit.edu.cn}
\and
\IEEEauthorblockN{2\textsuperscript{nd} Sendong Zhao}
\IEEEauthorblockA{\textit{Faculty of Computing,} \\
\textit{Harbin Institute of Technology}\\
Harbin, China \\
sdzhao@ir.hit.edu.cn}
\and
\IEEEauthorblockN{3\textsuperscript{rd} Haochun Wang}
\IEEEauthorblockA{\textit{Faculty of Computing,} \\
\textit{Harbin Institute of Technology}\\
Harbin, China \\
hcwang@ir.hit.edu.cn}
\and
\IEEEauthorblockN{4\textsuperscript{th} Yanrui Du}
\IEEEauthorblockA{\textit{Faculty of Computing,} \\
\textit{Harbin Institute of Technology}\\
Harbin, China \\
yrdu.hit@gmail.com}
\and
\IEEEauthorblockN{5\textsuperscript{th} Zewen Qiang}
\IEEEauthorblockA{\textit{Faculty of Computing,} \\
\textit{Harbin Institute of Technology}\\
Harbin, China \\
zwqiang@ir.hit.edu.cn}
\and
\IEEEauthorblockN{6\textsuperscript{th} Bing Qin}
\IEEEauthorblockA{\textit{Faculty of Computing,} \\
\textit{Harbin Institute of Technology}\\
Harbin, China \\
qinb@ir.hit.edu.cn}
\and
\IEEEauthorblockN{7\textsuperscript{th} Ting Liu}
\IEEEauthorblockA{\textit{Faculty of Computing,} \\
\textit{Harbin Institute of Technology}\\
Harbin, China \\
tliu@ir.hit.edu.cn}
}

\maketitle

\input{sections/0-abstract}
\input{sections/1-introduction} 
\input{sections/2-related_work} 
\input{sections/3-methods} 
\input{sections/4-experiments} 
\input{sections/5-conclusion} 

\renewcommand{\refname}{}
\section{References}\label{sec:reference}

\bibliographystyle{IEEEtran}
\bibliography{IEEEexample}

\end{document}

%% file: sections/0-abstract.tex
\begin{abstract}

Artificial Intelligence predicts drug properties by encoding drug molecules, aiding in the rapid screening of candidates. Different molecular representations, such as SMILES and molecule graphs, contain complementary information for molecular encoding. Thus exploiting complementary information from different molecular representations is one of the research priorities in molecular encoding. 
Most existing methods for combining molecular multi-modalities only use molecular-level information, making it hard to encode intra-molecular alignment information between different modalities. 
To address this issue, we propose a multi-granularity fusion method that is MolFusion. The proposed MolFusion consists of two key components: (1) MolSim, a molecular-level encoding component that achieves molecular-level alignment between different molecular representations.
and (2) AtomAlign, an atomic-level encoding component that achieves atomic-level alignment between different molecular representations.
Experimental results show that MolFusion effectively utilizes complementary multimodal information, leading to significant improvements in performance across various classification and regression tasks.

\end{abstract}

\begin{IEEEkeywords}
multimodality, fusion method, multi-granularity views
\end{IEEEkeywords}

%% file: sections/1-introduction.tex
\section{Introduction}


Artificial Intelligence (AI) enhances drug property prediction by encoding molecular structures, thus facilitating the rapid identification of unqualified drug candidates from extensive pools \cite{xia2022systematic, gawehn2016deep, goh2017deep, zeng2022deep, su2022molecular, payne2020bert, zhang2021motif}. Studies \cite{zhu2021dual, hou2022graphmae, zhu2022featurizations} indicate that different molecular representations capture distinct and complementary aspects of molecular information, highlighting the potential benefits of integrating multiple modalities to fully understand and utilize the diverse characteristics of molecules.

\begin{figure}[h]
    \centering
    \includegraphics[width=0.48\textwidth]{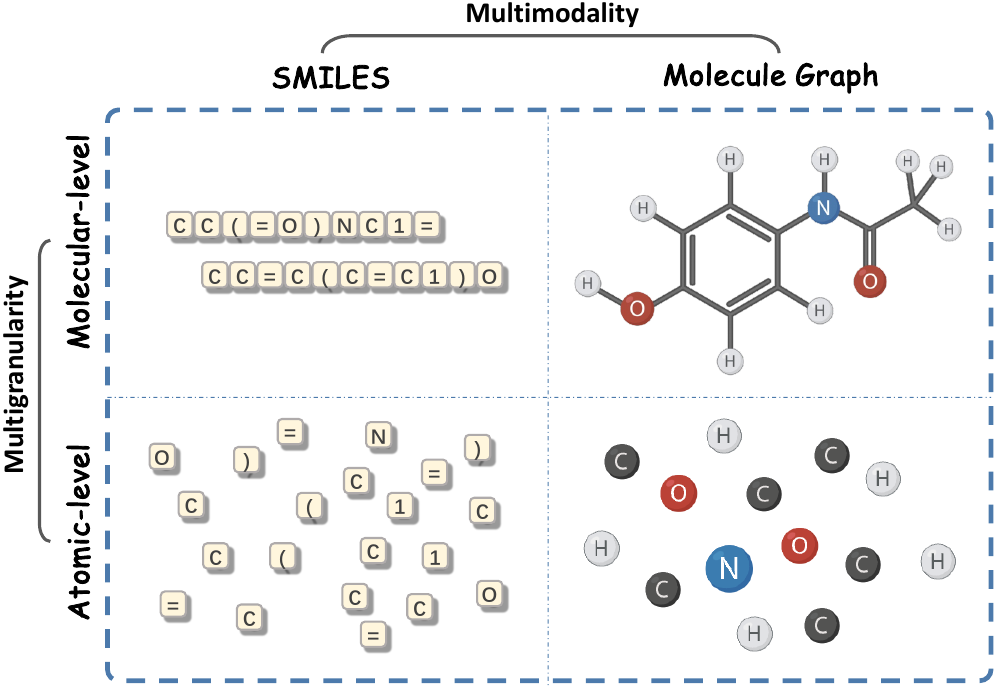}
    \caption{Diagram of multimodality and multi-granularity. It illustrates how both SMILES and molecule graphs can be represented at molecular and atomic levels, providing a comprehensive view of different granularities and modalities.}
\label{multi}
\end{figure}

It is reasonable to assume that different molecular representations of the same molecule possess complementary information because they encode distinct underlying knowledge (\textbf{Complementary Information Assumption}).
Therefore, integrating complementary information from different molecular representations becomes a key research direction in molecular encoding. However, most existing studies focus only on combining molecular-level information and fail to incorporate intra-molecular alignment information between different modalities. For example, DMP \cite{zhu2021dual}, PanGu \cite{lin2022pangu}, GraphMVP \cite{hou2022graphmae}, and MEMO \cite{zhu2022featurizations} rely on contrastive learning or self-reconstruction techniques to align different representations. Consequently, much of the complementary information between different representations is not effectively utilized during molecular-level alignment.

To address the above issue, we consider both molecular-level and atomic-level alignments of different molecular representations to fully utilize complementary information (see Figure \ref{multi}). Specifically, we propose MolFusion, a multi-granularity fusion learning method that 1) integrates different molecular representations (e.g., SMILES and molecule graphs); and 2) exploits both molecular-level and atomic-level alignments. The proposed molecular-level component, \textbf{MolSim}, brings the encoded representations of similar molecules closer together in continuous vector space by enhancing contrastive learning with similarity scores derived from molecular knowledge, replacing traditional binary labels. Our atomic-level component, \textbf{AtomAlign}, enriches atomic-level encoding by using one representation to complete the masked atoms in the other representation of the same molecule. In detail, it innovatively predicts masked information in SMILES by leveraging the encoded representations of masked SMILES and unmasked molecule graphs. By synchronously training these two components, we achieve partial alignment of the encoded representations of the two molecular modalities, thereby significantly preserving complementary information.



To validate the effectiveness of our proposed multi-granularity fusion learning method, we selected 6 classification tasks and 3 regression tasks from MoleculeNet \cite{wu2018moleculenet} for evaluation. The experimental results demonstrate that our method achieves significant performance improvements compared to other fusion methods. Furthermore, ablation study and visualization confirm the validity of our multi-granularity fusion approach, highlighting its robustness in preserving complementary information between different molecular modalities. Therefore, our proposed fusion learning method surpasses existing fusion techniques and lays a solid foundation for future research in multimodal molecular representation learning. Additionally, our approach is model-agnostic, allowing for the integration of any existing pre-trained powerful models.

In our research, we propose a multimodal and multi-granularity fusion learning. Our contributions are as follows:
\begin{itemize}

    \item \textbf{Multi-granularity views}: We introduce an innovative atomic-level component that enriches atomic-level encoding by using one representation to complete the masked atoms in the other representation of the same molecule. Additionally, we propose a molecular-level component that brings the encoded representations of similar molecules closer together in continuous vector space.

    \item \textbf{Mutimodal confusion}: Our method effectively leverages complementary information from different molecular modalities. By integrating SMILES and molecular graphs, we harness the unique strengths of each representation to enhance the overall molecular encoding, ensuring robust preservation of complementary information.

    \item \textbf{Effective Strategy}: The results show that our integrated model, despite its simplicity, outperforms existing fusion methods, demonstrating the effectiveness of our multimodal fusion strategy in enhancing molecular representation learning.
\end{itemize}

%% file: sections/2-related_work.tex
\section{Related Work}
In this section, we analyze the development and shortcomings of current molecular representation methods, focusing on the perspective of molecular data modality.


\subsection{Unimodal Molecular Representation Methods}



Numerous molecular representation learning methods are based on single molecular representations, such as SMILES \cite{weininger1988smiles}, molecular graphs \cite{li2022deep}, and 3D molecular representations \cite{li2022deep}. There are relatively comprehensive methods for both molecular-level and atomic-level training for unimodal molecular representation learning. In molecular-level methods, contrastive learning \cite{radford2021learning, fang2023knowledge, furst2022cloob} is a common training technique. MolCLR \cite{wang2102molclr}, MoCL \cite{sun2021mocl}, and GraphCL \cite{you2020graph} perform contrastive learning on augmented molecular graphs after data augmentation operations. Knowledge-based BERT \cite{wu2022knowledge} generates various SMILES from canonical SMILES for data augmentation. In MolGNet \cite{li2021effective}, molecular graphs are split into two halves. Two halves from the same molecule are used as positive samples, and halves from different molecules are used as negative samples. KCL \cite{fang2022molecular} enhances molecular graphs using knowledge graphs, adding corresponding attribute nodes to atoms and connecting them with edges. The molecular graphs before and after knowledge enhancement are encoded separately, with the encodings of the same molecule as positive samples and the encodings of different molecules as negative samples.

Beyond contrastive learning, SMILES Transformer \cite{wang2019smiles} uses the autoencoding method to learn molecular representations. Similarly, X-Mol \cite{xue2020x} designs Transformer layers, using bidirectional self-attention layers in the first half to learn SMILES representations and unidirectional self-attention layers in the second half to regenerate SMILES from the learned embeddings. Grover \cite{rong2020self} and knowledge-based BERT propose a molecular features prediction task, predicting molecular features, such as the presence of specific functional groups, from the embeddings of the entire molecule.

For atomic-level methods, the most classic approach is masked language models \cite{honda2019smiles, chithrananda2020chemberta, zhang2021mg}. CHEM-BERT \cite{kim2021merged}, for example, masks atoms in SMILES and uses molecular contextual information to predict the types of masked atoms. For molecular graphs, atom features are masked, and other molecular features, such as bond features, are used to predict the types of atoms \cite{hu2019strategies, li2021effective, li2023knowledge, xia2022mole}. 

Beyond atom masking tasks, Grover proposes using the learned atomic features to predict the contextual properties of atoms. In the context prediction task \cite{hu2019strategies}, atom neighborhood encodings are used as node embeddings, and surrounding graph structures are represented as context embeddings. The task is to predict whether the node embeddings and context embeddings come from the same molecule. InfoGraph \cite{sun2019infograph} proposes to predict which atoms belong to which molecules based on representations of all molecules and the atoms they contain.

Notably, Hu et al. \cite{hu2019strategies} suggests strategies for pre-training at the molecular or atomic level are found to offer limited improvements and may even degrade the performance of many downstream tasks. Therefore, most works \cite{jiang2022multigran, rong2020self, li2021effective} attempt to retain both molecular-level and atomic-level training strategies in their studies, employing multi-granularity methods to ensure optimal performance.


\subsection{Multimodal Molecular Representation Methods}
Researchers \cite{zhu2021dual, hou2022graphmae, zhu2022featurizations} find that different molecular representations can provide complementary information, making the effective utilization of multiple molecular representations a key focus in drug property prediction. For instance, DMP \cite{zhu2021dual} and PanGu \cite{lin2022pangu} utilize molecular linear representations and molecular graph information, GraphMVP \cite{hou2022graphmae} leverages molecular graphs and 3D molecular representations, while MEMO \cite{zhu2022featurizations} processes multiple representations, including SMILES, molecular graphs, 3D representations, and fingerprints. In multimodal methods, contrastive learning and its variants are prevalent. For example, GraphMVP and DMP treat different representations of the same molecule as positive samples and representations from different molecules as negative samples for contrastive learning or its variants. MEMO treats single representations and aggregated representations from the same molecule as positive samples and those from different molecules as negative samples. 

Beyond contrastive learning, self-reconstructing is another approach. GraphMVP reconstructs one representation from the other between molecular graphs and 3D molecular representations. Similarly, PanGu uses molecular graphs to reconstruct the linear representation of SELFIES \cite{krenn2020self, janakarajan2023language}, which is similar to SMILES.


Unfortunately, these works lack exploration of multimodal atomic-level approaches and also ignore the use of existing trained, powerful, unimodal models. For instance, DMP is a typical work that integrates SMILES and molecule graphs, using atomic-level masking methods and dual-view molecular-level contrastive learning on two single modalities. However, DMP trains two encoders from scratch, neglecting the use of the current powerful single modal encoders, resulting in the need for a large amount of data: 110M of training data, which is about 4.4 times our training data and consumes significant computational resources.
\\

Therefore, our approach aims to propose a multi-granularity molecular representation fusion learning, utilizing atomic-level alignment to enhance the complementary information between different molecular representations.

%% file: sections/3-methods.tex
\section{Methodology}
In this section, we introduce the molecular-level component MolSim and the atomic-level component AtomAlign to achieve fusion learning of pre-trained encoders for different molecular representations. Molecular-level/atomic-level methods (see Figure \ref{multi}) represent the methods with molecules/atoms as the minimum processing units, respectively.

\subsection{Molecular-level Fusion Component: MolSim}

\begin{figure}[H] 
    \centering
    \includegraphics[width=0.49\textwidth]{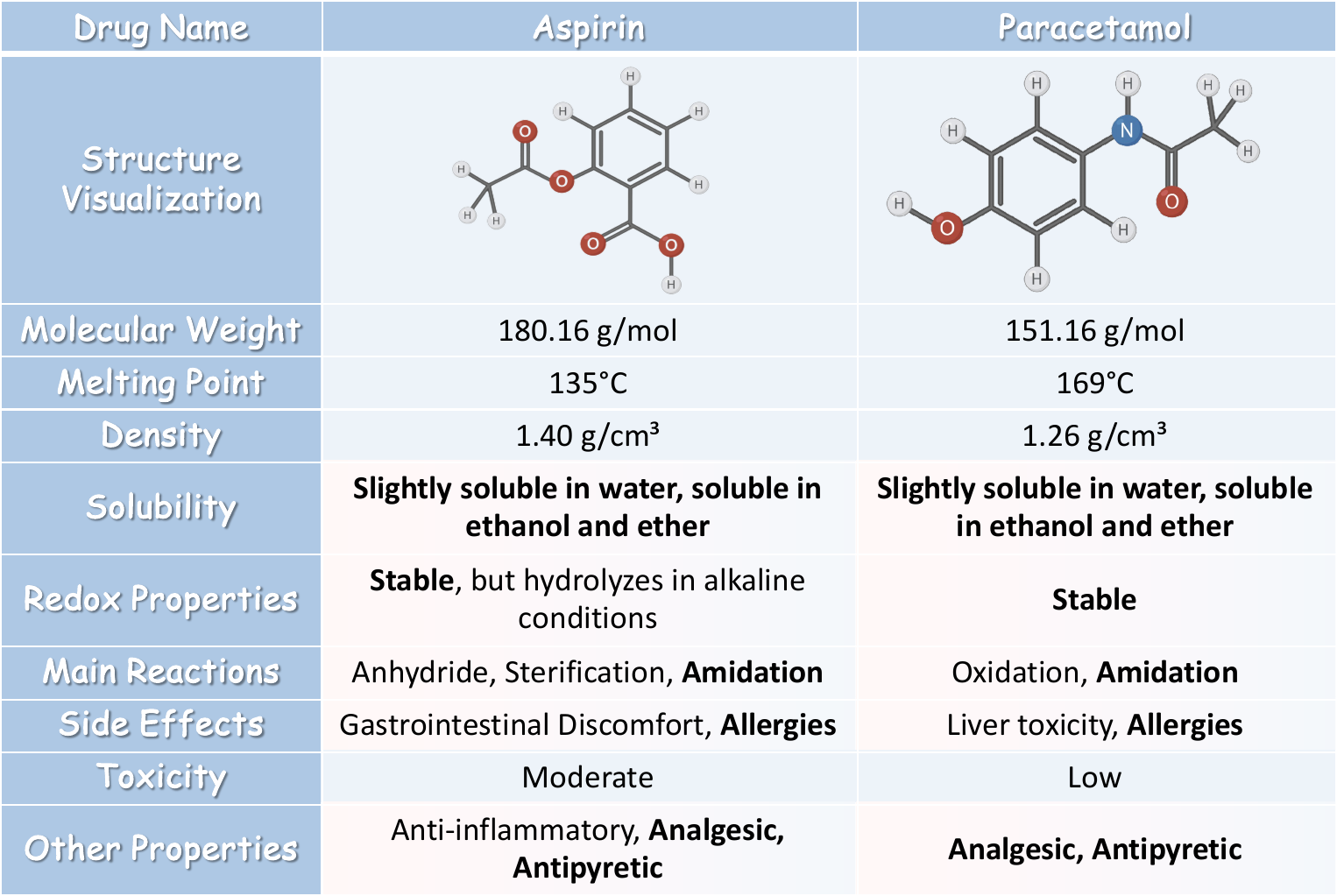}
    \caption{Diagram of molecular similarity. Comparison of the molecular properties of Aspirin and Paracetamol, with similar properties highlighted in bold.}
    \label{sim}
\end{figure}

Contrastive learning \cite{radford2021learning} is a typically molecular-level method for multimodal fusion, labeling different representations of the same molecule as 1 and those of different molecules as 0. This approach implies that any two different molecules are entirely dissimilar \cite{xia2021progcl, wang2022medclip}. However, as shown in Figure \ref{sim}, although Aspirin and Paracetamol are different molecules, they share similar properties, such as solubility in water and ethanol and analgesic effects. To better capture molecular similarities, we introduce MolSim, which uses continuous similarity measures instead of traditional binary labels. MolSim thus captures subtle molecular relationships more accurately.

For accurate similarity measurement, we utilize Morgan fingerprints \cite{rogers2010extended} to compute the Tanimoto coefficient \cite{vogt2017modeling}, which captures detailed molecular information, including atom types, bond types, and functional groups. The coefficient effectively reflects structural similarity, providing a reasonable measure of molecular relationships.

\begin{figure*}[h]
    \centering
    \includegraphics[width=\textwidth]{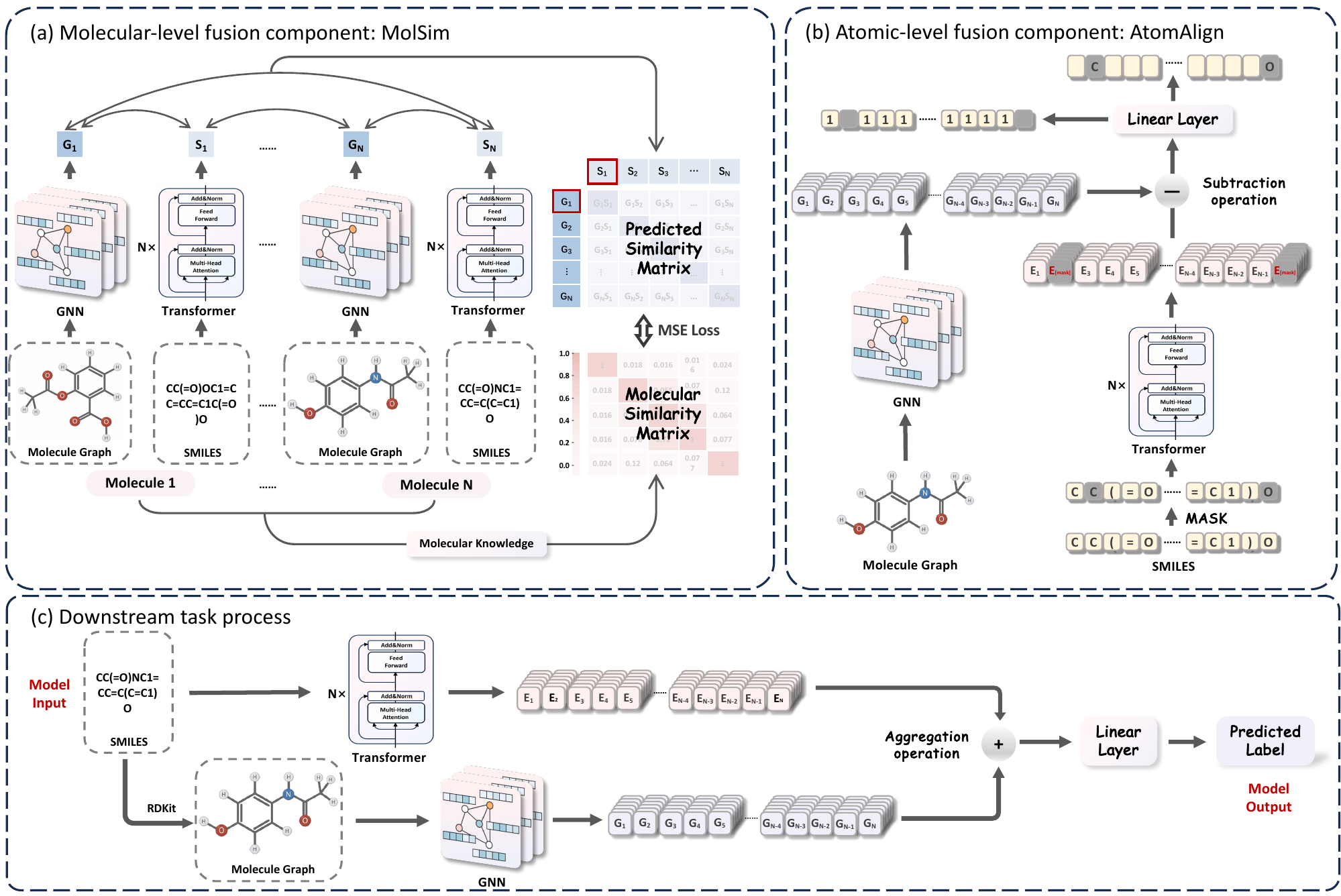}
    \caption{Overview of MolFusion. 
        (a) \textbf{MolSim component Illustration}: Molecule graph and SMILES representations are processed through existing GNN and Transformer encoders, respectively. The computed similarity matrix is compared with the molecular similarity matrix using MSE loss.
        (b) \textbf{AtomAlign component Illustration}: Randomly masked SMILES are encoded and subtracted from the molecular graph encodings. The resulting difference vector is used to predict masked information by only introducing a linear layer. MolSim and AtomAlign are trained synchronously.
        (c) \textbf{Downstream Task Process}: SMILES and molecular graphs are encoded by their respective encoders trained through fusion learning. The resulting encodings are aggregated and then passed through a linear layer to predict the outcome. In all downstream tasks, the parameters of both encoders are frozen. Due to the accessibility of SMILES and molecular graph data, either input can be used to generate the other using the RDKit tool \cite{landrum2013rdkit}, thereby leveraging both fusion-learned encoders. The aggregation operation includes encoder-only, element-wise addition, and concatenation operations.
}
    \label{F_L}
\end{figure*}

In the MolSim component, we replace binary labels with continuous similarity measures by calculating the Tanimoto coefficient between Morgan fingerprints, as shown in Figure \ref{F_L} (a). This continuous similarity measure allows MolSim to learn more effectively and leverage complementary information.

The similarity matrix \( T \), calculated by using the Tanimoto coefficient between Morgan fingerprints, as mentioned above, serves as the target labels for learning. Given the features from SMILES and molecule graphs, the computed similarity matrix \( Q \) is computed as:
\begin{equation}
Q = \tau^{-1} \times S \times G^T
\end{equation}

where \( \tau^{-1} \) is the inverse of the temperature parameter \( \tau \), and \( S \) and \( G \) represent the feature matrices for SMILES and molecule graphs, respectively.

In MolSim, we use the mean squared error to calculate the loss between model predictions and the molecular similarity matrix \( T \):
\begin{equation}
\mathcal{L}_{\text{MolSim}} = \frac{1}{N \times N} \sum_{i=1}^{N} \sum_{j=1}^{N} (Q_{ij} - T_{ij})^2
\end{equation}

where \( N \) is the number of samples in the matrix.

\subsection{Atomic-level Fusion Component: AtomAlign}
In addition to learning molecular-level representations, we introduce AtomAlign for atomic-level fusion. Through atomic alignment, the model can align identical information contained in different molecular representations and obtain unique molecular information from each representation. This is consistent with the \textbf{Complementary Information Assumption}. In AtomAlign, we employ atomic-level masking to achieve alignment of atomic features between different molecular representations, as illustrated in Figure \ref{F_L} (b).

In AtomAlign, we randomly mask atoms in SMILES and encode the masked SMILES. We then subtract the masked SMILES encodings from the molecular graph encodings of the same molecule. The resulting encodings are passed through a linear layer to predict both the types of masked atoms in SMILES and identify which atoms are not masked without predicting their types.

To align atomic features between different molecular representations, we first calculate the difference between the molecular graph encodings \(\mathcal{T}\) and the masked SMILES encodings \(\mathcal{E}_{\text{mask}}\). The resulting difference vector \(\mathcal{D}\) is given by:
\begin{equation}
\mathcal{D} = \mathcal{T} - \mathcal{E}_{\text{mask}}
\end{equation}

Our loss function integrates the prediction of masked atoms and the identification of unmasked atoms. The total loss is computed as follows:
\begin{equation}
\mathcal{L}_{\text{mask}} = \frac{1}{N_m} \sum_{i \in \text{masked}} \text{CrossEntropy}(\mathcal{D}_i, \mathcal{I}_i^{\text{mask}})
\end{equation}

\begin{equation}
\mathcal{L}_{\text{unmask}} = \frac{1}{N_u} \sum_{i \in \text{unmasked}} \text{CrossEntropy}(\mathcal{D}_{i}, \mathcal{M}_{i}^{\text{unmasked}})
\end{equation}
\begin{equation}
\mathcal{L}_{\text{AtomAlign}} = \alpha \mathcal{L}_{\text{mask}} + (1 - \alpha) \mathcal{L}_{\text{unmask}}
\end{equation}

where \( \mathcal{D}_i \) represents the difference vector at the \( i^{\text{th}} \) position, \( \mathcal{I}_i^{\text{mask}} \) is the index label of the masked atom at position \( i \) in the SMILES representation. \( \mathcal{M} \) represents the masking indicator matrix, where \( \mathcal{M}_i \) is the 0/1 label indicating whether the atom at the \( i^{\text{th}} \) position is masked, and \( \alpha \) is a weighting parameter that balances the contributions of the masked and unmasked loss components, with \( 0 \leq \alpha \leq 1 \).

These two components are trained synchronously, therefore:
\begin{equation}
\mathcal{L}_{\text{MolFusion}} = \mathcal{L}_{\text{MolSim}} + \beta \cdot \mathcal{L}_{\text{AtomAlign}}
\end{equation}

where \(\beta\) is a hyperparameter.
\\


Our method design offers several advantages: (1) Integrating molecular-level and atomic-level fusion learning effectively leverages the complementary information from different molecular representations. (2) Our method significantly reduces data and computational resource consumption by efficiently using the existing powerful unimodal encoders. (3) Due to the accessibility of SMILES and molecular graph data, even if only SMILES or molecular graph is provided as input in downstream tasks, we can use the RDKit tool \cite{landrum2013rdkit} to obtain another modal representation of molecules, thereby jointly improving model performance, as shown in Figure \ref{F_L} (c). (4) Our design is notably lightweight, introducing only a single linear layer in the AtomAlign task, which avoids the inclusion of excessive model parameters and reduces training costs.

%% file: sections/4-experiments.tex
\section{Experiments}
In this section, we introduce experimental design, including the backbones, datasets, and baselines. Besides, we validate the effectiveness of our approach using 6 classification tasks and 3 regression tasks from MoleculeNet \cite{wu2018moleculenet}. Additionally, we verify the \textbf{Complementary Information Assumption} and demonstrate the rationality of the components through comprehensive experimental analysis.

\subsection{Experimental Design}
\subsubsection{Backbones}

\textbf{SMILES Encoder}: We select CHEM-BERT \cite{kim2021merged} as the SMILES encoder in our study because CHEM-BERT demonstrates robust performance in drug property prediction. Its ability to encode SMILES into continuous embeddings effectively captures unique molecular characteristics.

\textbf{Molecular Graph Encoder}: For molecular graph encoders, we use the Grover-large model \cite{rong2020self}. Grover-large is a powerful graph neural network-based encoder that can effectively process and learn representations of molecular graphs.


\begin{table*}[htbp]
\caption{Comparison of performance of various fusion methods and aggregation operations on different datasets. The second row represents the number of samples in each dataset. In the ``Fusion Method'' column, ``NO-TRAIN'' (corresponding to section \ref{simple_fusion}) indicates no additional training for the two encoders, and ``CL'' stands for contrastive learning. In the ``Aggregation Operation'' column, ``MG'' represents Molecule Graph Encoder-only, ``smiles'' represents SMILES Encoder-only, and CCO and EWA represent two proposed aggregation operations. For the NO-TRAIN and SMILES Encoder-only condition, the performance reflects that of CHEM-BERT, while for the No-Train and Molecule Graph Encoder-only condition, it reflects that of GROVER-LARGE. In downstream experiments, we freeze the parameters of the two encoders and use a linear probe for prediction. The best performance in each dataset is highlighted in bold.}
\begin{center}
\label{tab:performance_comparison}
\renewcommand{\arraystretch}{1.2}
\begin{tabular}{p{0.4cm} p{1cm} |c c c c c c c c c c c c |c c c c c c}
\hline
                              &                           & \multicolumn{2}{c}{SIDER}            & \multicolumn{2}{c}{BBBP}             & \multicolumn{2}{c}{BACE}              & \multicolumn{2}{c}{Clintox}          & \multicolumn{2}{c}{Tox21}            & \multicolumn{2}{c}{Toxcast}          & \multicolumn{2}{|c}{ESOL}              & \multicolumn{2}{c}{Lipo}              & \multicolumn{2}{c}{Freesolv}          \\ \cline{3-20} 
                              &                           & \multicolumn{2}{c}{1427}             & \multicolumn{2}{c}{2050}             & \multicolumn{2}{c}{1514}              & \multicolumn{2}{c}{1484}             & \multicolumn{2}{c}{7831}             & \multicolumn{2}{c}{8597}             & \multicolumn{2}{|c}{1128}              & \multicolumn{2}{c}{4201}              & \multicolumn{2}{c}{642}               \\ \cline{3-20}
\multirow{-3}{*}{\parbox{1.2cm}{\scriptsize Fusion \newline Method}}
        & \multirow{-3}{*}{\parbox{1.1cm}{\fontsize{6.5}{8}\selectfont Aggregation \newline Operation}}
        & \multicolumn{12}{c}{ROC-AUC}                                                                                                                                                                                                              & \multicolumn{6}{|c}{RMSE}                                                                                              \\ \cline{1-20}
                              & MG                        & \cellcolor[HTML]{86C97E}\makebox[0.6cm][c]{55.64} & \makebox[0.3cm][c]{3.04} & \cellcolor[HTML]{FEE582}\makebox[0.6cm][c]{55.77} & \makebox[0.3cm][c]{1.19} & \cellcolor[HTML]{F8756D}\makebox[0.6cm][c]{38.35} & \makebox[0.3cm][c]{2.44}  & \cellcolor[HTML]{FDCB7D}\makebox[0.6cm][c]{47.13} & \makebox[0.3cm][c]{3.81} & \cellcolor[HTML]{FCEB84}\makebox[0.6cm][c]{52.05} & \makebox[0.3cm][c]{0.34} & \cellcolor[HTML]{F8746D}\makebox[0.6cm][c]{48.83} & \makebox[0.3cm][c]{0.31} & \cellcolor[HTML]{DDE182}\makebox[0.6cm][c]{2.071} & \makebox[0.3cm][c]{0.077} & \cellcolor[HTML]{F1E783}\makebox[0.6cm][c]{1.039} & \makebox[0.3cm][c]{0.018} & \cellcolor[HTML]{FFDC82}\makebox[0.6cm][c]{4.197} & \makebox[0.3cm][c]{0.051} \\
                              & SMILES                    & \cellcolor[HTML]{F98470}\makebox[0.6cm][c]{50.21} & \makebox[0.3cm][c]{0.11} & \cellcolor[HTML]{88C97E}\makebox[0.6cm][c]{60.01} & \makebox[0.3cm][c]{0.35} & \cellcolor[HTML]{BFD981}\makebox[0.6cm][c]{50.47} & \makebox[0.3cm][c]{9.25}  & \cellcolor[HTML]{F8696B}\makebox[0.6cm][c]{39.47} & \makebox[0.3cm][c]{0.65} & \cellcolor[HTML]{C2DA81}\makebox[0.6cm][c]{56.38} & \makebox[0.3cm][c]{0.06} & \cellcolor[HTML]{90CB7E}\makebox[0.6cm][c]{51.31} & \makebox[0.3cm][c]{0.06} & \cellcolor[HTML]{F8696B}\makebox[0.6cm][c]{2.256} & \makebox[0.3cm][c]{0.006} & \cellcolor[HTML]{FA8771}\makebox[0.6cm][c]{1.100}   & \makebox[0.3cm][c]{0.008} & \cellcolor[HTML]{D7DF81}\makebox[0.6cm][c]{4.163} & \makebox[0.3cm][c]{0.037} \\ 
                              & CCO              & \cellcolor[HTML]{DAE182}\makebox[0.6cm][c]{54.56} & \makebox[0.3cm][c]{2.15} & \cellcolor[HTML]{B9D780}\makebox[0.6cm][c]{58.41} & \makebox[0.3cm][c]{0.28} & \cellcolor[HTML]{F8696B}\makebox[0.6cm][c]{37.59} & \makebox[0.3cm][c]{7.85}  & \cellcolor[HTML]{FCB87A}\makebox[0.6cm][c]{45.61} & \makebox[0.3cm][c]{3.83} & \cellcolor[HTML]{E5E483}\makebox[0.6cm][c]{53.77} & \makebox[0.3cm][c]{0.35} & \cellcolor[HTML]{FBA576}\makebox[0.6cm][c]{49.25} & \makebox[0.3cm][c]{0.37} & \cellcolor[HTML]{DAE081}\makebox[0.6cm][c]{2.065} & \makebox[0.3cm][c]{0.140}  & \cellcolor[HTML]{93CC7D}\makebox[0.6cm][c]{0.989} & \makebox[0.3cm][c]{0.020}  & \cellcolor[HTML]{FDC27C}\makebox[0.6cm][c]{4.203} & \makebox[0.3cm][c]{0.042} \\
\multirow{-4}{*}{\parbox{2.5cm}{No- \newline Train}}        & EWA                & \cellcolor[HTML]{70C27C}\makebox[0.6cm][c]{55.93} & \makebox[0.3cm][c]{3.21} & \cellcolor[HTML]{F6E984}\makebox[0.6cm][c]{56.41} & \makebox[0.3cm][c]{0.95} & \cellcolor[HTML]{F98971}\makebox[0.6cm][c]{39.49} & \makebox[0.3cm][c]{3.69}  & \cellcolor[HTML]{FEEB84}\makebox[0.6cm][c]{49.86} & \makebox[0.3cm][c]{3.79} & \cellcolor[HTML]{DEE283}\makebox[0.6cm][c]{54.32} & \makebox[0.3cm][c]{0.98} & \cellcolor[HTML]{FA9673}\makebox[0.6cm][c]{49.12} & \makebox[0.3cm][c]{0.31} & \cellcolor[HTML]{FFEA84}\makebox[0.6cm][c]{2.141} & \makebox[0.3cm][c]{0.119} & \cellcolor[HTML]{E0E282}\makebox[0.6cm][c]{1.030}  & \makebox[0.3cm][c]{0.019} & \cellcolor[HTML]{CADB80}\makebox[0.6cm][c]{4.153} & \makebox[0.3cm][c]{0.030}  \\ \cline{1-20}
                              & MG                        & \cellcolor[HTML]{FEDC81}\makebox[0.6cm][c]{53.54} & \makebox[0.3cm][c]{2.28} & \cellcolor[HTML]{FDCE7E}\makebox[0.6cm][c]{54.37} & \makebox[0.3cm][c]{2.00} & \cellcolor[HTML]{FDD47F}\makebox[0.6cm][c]{43.89} & \makebox[0.3cm][c]{1.52}  & \cellcolor[HTML]{A7D27F}\makebox[0.6cm][c]{63.33} & \makebox[0.3cm][c]{2.47} & \cellcolor[HTML]{FCBA7A}\makebox[0.6cm][c]{50.27} & \makebox[0.3cm][c]{0.13} & \cellcolor[HTML]{F8696B}\makebox[0.6cm][c]{48.73} & \makebox[0.3cm][c]{0.17} & \cellcolor[HTML]{FCAF79}\makebox[0.6cm][c]{2.194} & \makebox[0.3cm][c]{0.035} & \cellcolor[HTML]{FFDD82}\makebox[0.6cm][c]{1.054} & \makebox[0.3cm][c]{0.007} & \cellcolor[HTML]{FDBD7C}\makebox[0.6cm][c]{4.204} & \makebox[0.3cm][c]{0.009} \\
                              & SMILES                    & \cellcolor[HTML]{F4E884}\makebox[0.6cm][c]{54.23} & \makebox[0.3cm][c]{2.34} & \cellcolor[HTML]{FEDD81}\makebox[0.6cm][c]{55.28} & \makebox[0.3cm][c]{2.38} & \cellcolor[HTML]{EDE683}\makebox[0.6cm][c]{46.70} & \makebox[0.3cm][c]{7.46}  & \cellcolor[HTML]{F86C6B}\makebox[0.6cm][c]{39.71} & \makebox[0.3cm][c]{1.15} & \cellcolor[HTML]{FEDC81}\makebox[0.6cm][c]{51.35} & \makebox[0.3cm][c]{0.06} & \cellcolor[HTML]{E2E383}\makebox[0.6cm][c]{50.23} & \makebox[0.3cm][c]{0.03} & \cellcolor[HTML]{FCAE79}\makebox[0.6cm][c]{2.195} & \makebox[0.3cm][c]{0.040}  & \cellcolor[HTML]{FB9474}\makebox[0.6cm][c]{1.093} & \makebox[0.3cm][c]{0.011} & \cellcolor[HTML]{EDE582}\makebox[0.6cm][c]{4.180}  & \makebox[0.3cm][c]{0.024} \\
                              & CCO              & \cellcolor[HTML]{FEDD81}\makebox[0.6cm][c]{53.56} & \makebox[0.3cm][c]{2.20}  & \cellcolor[HTML]{BFD981}\makebox[0.6cm][c]{58.21} & \makebox[0.3cm][c]{2.63} & \cellcolor[HTML]{BCD881}\makebox[0.6cm][c]{50.77} & \makebox[0.3cm][c]{3.29}  & \cellcolor[HTML]{FCC47C}\makebox[0.6cm][c]{46.54} & \makebox[0.3cm][c]{0.37} & \cellcolor[HTML]{FEE382}\makebox[0.6cm][c]{51.59} & \makebox[0.3cm][c]{0.41} & \cellcolor[HTML]{FCBE7B}\makebox[0.6cm][c]{49.47} & \makebox[0.3cm][c]{0.08} & \cellcolor[HTML]{F9E983}\makebox[0.6cm][c]{2.129} & \makebox[0.3cm][c]{0.081} & \cellcolor[HTML]{CEDC81}\makebox[0.6cm][c]{1.020}  & \makebox[0.3cm][c]{0.015} & \cellcolor[HTML]{F8696B}\makebox[0.6cm][c]{4.223} & \makebox[0.3cm][c]{0.022} \\
\multirow{-4}{*}{CL}          & EWA                & \cellcolor[HTML]{FEE783}\makebox[0.6cm][c]{53.93} & \makebox[0.3cm][c]{2.21} & \cellcolor[HTML]{FEDC81}\makebox[0.6cm][c]{55.23} & \makebox[0.3cm][c]{3.25} & \cellcolor[HTML]{FEE582}\makebox[0.6cm][c]{44.88} & \makebox[0.3cm][c]{5.92}  & \cellcolor[HTML]{FEE582}\makebox[0.6cm][c]{49.15} & \makebox[0.3cm][c]{1.77} & \cellcolor[HTML]{FCC27C}\makebox[0.6cm][c]{50.51} & \makebox[0.3cm][c]{0.05} & \cellcolor[HTML]{FEDF81}\makebox[0.6cm][c]{49.75} & \makebox[0.3cm][c]{0.05} & \cellcolor[HTML]{D4DE81}\makebox[0.6cm][c]{2.054} & \makebox[0.3cm][c]{0.087} & \cellcolor[HTML]{FFE082}\makebox[0.6cm][c]{1.052} & \makebox[0.3cm][c]{0.002} & \cellcolor[HTML]{FDB97B}\makebox[0.6cm][c]{4.205} & \makebox[0.3cm][c]{0.013} \\ \cline{1-20}
                              & MG                        & \cellcolor[HTML]{F8696B}\makebox[0.6cm][c]{49.18} & \makebox[0.3cm][c]{0.72} & \cellcolor[HTML]{FCB679}\makebox[0.6cm][c]{52.89} & \makebox[0.3cm][c]{6.55} & \cellcolor[HTML]{63BE7B}\makebox[0.6cm][c]{\textbf{57.99}} & \makebox[0.3cm][c]{10.47} & \cellcolor[HTML]{F6E984}\makebox[0.6cm][c]{50.99} & \makebox[0.3cm][c]{4.31} & \cellcolor[HTML]{F98370}\makebox[0.6cm][c]{48.50}  & \makebox[0.3cm][c]{0.42} & \cellcolor[HTML]{E3E383}\makebox[0.6cm][c]{50.22} & \makebox[0.3cm][c]{0.24} & \cellcolor[HTML]{F96F6D}\makebox[0.6cm][c]{2.251} & \makebox[0.3cm][c]{0.150}  & \cellcolor[HTML]{F96F6D}\makebox[0.6cm][c]{1.113} & \makebox[0.3cm][c]{0.004} & \cellcolor[HTML]{FFE583}\makebox[0.6cm][c]{4.195} & \makebox[0.3cm][c]{0.041} \\
                              & SMILES                    & \cellcolor[HTML]{FBA075}\makebox[0.6cm][c]{51.29} & \makebox[0.3cm][c]{1.21} & \cellcolor[HTML]{FA9D75}\makebox[0.6cm][c]{51.38} & \makebox[0.3cm][c]{1.53} & \cellcolor[HTML]{FA9E75}\makebox[0.6cm][c]{40.74} & \makebox[0.3cm][c]{4.70}   & \cellcolor[HTML]{F6E984}\makebox[0.6cm][c]{51.09} & \makebox[0.3cm][c]{3.61} & \cellcolor[HTML]{FA9C74}\makebox[0.6cm][c]{49.31} & \makebox[0.3cm][c]{1.44} & \cellcolor[HTML]{FEEB84}\makebox[0.6cm][c]{49.87} & \makebox[0.3cm][c]{0.41} & \cellcolor[HTML]{F9706D}\makebox[0.6cm][c]{2.250}  & \makebox[0.3cm][c]{0.013} & \cellcolor[HTML]{F96F6D}\makebox[0.6cm][c]{1.113} & \makebox[0.3cm][c]{0.004} & \cellcolor[HTML]{FDEA83}\makebox[0.6cm][c]{4.192} & \makebox[0.3cm][c]{0.033} \\
                              & CCO              & \cellcolor[HTML]{F86E6C}\makebox[0.6cm][c]{49.40}  & \makebox[0.3cm][c]{1.81} & \cellcolor[HTML]{F8696B}\makebox[0.6cm][c]{48.14} & \makebox[0.3cm][c]{4.53} & \cellcolor[HTML]{F98D72}\makebox[0.6cm][c]{39.75} & \makebox[0.3cm][c]{1.80}   & \cellcolor[HTML]{FDD47F}\makebox[0.6cm][c]{47.79} & \makebox[0.3cm][c]{5.45} & \cellcolor[HTML]{F97F6F}\makebox[0.6cm][c]{48.38} & \makebox[0.3cm][c]{0.60}  & \cellcolor[HTML]{DCE182}\makebox[0.6cm][c]{50.32} & \makebox[0.3cm][c]{0.15} & \cellcolor[HTML]{F9766E}\makebox[0.6cm][c]{2.245} & \makebox[0.3cm][c]{0.006} & \cellcolor[HTML]{F8696B}\makebox[0.6cm][c]{1.116} & \makebox[0.3cm][c]{0.003} & \cellcolor[HTML]{EAE582}\makebox[0.6cm][c]{4.178} & \makebox[0.3cm][c]{0.017} \\
\multirow{-4}{*}{DMP}         & EWA                & \cellcolor[HTML]{F86E6C}\makebox[0.6cm][c]{49.40}  & \makebox[0.3cm][c]{1.74} & \cellcolor[HTML]{FCB87A}\makebox[0.6cm][c]{53.01} & \makebox[0.3cm][c]{6.62} & \cellcolor[HTML]{66BF7C}\makebox[0.6cm][c]{57.79} & \makebox[0.3cm][c]{10.4}  & \cellcolor[HTML]{EFE784}\makebox[0.6cm][c]{52.07} & \makebox[0.3cm][c]{3.35} & \cellcolor[HTML]{F8696B}\makebox[0.6cm][c]{47.66} & \makebox[0.3cm][c]{0.82} & \cellcolor[HTML]{FEE883}\makebox[0.6cm][c]{49.83} & \makebox[0.3cm][c]{0.25} & \cellcolor[HTML]{F8696B}\makebox[0.6cm][c]{2.256} & \makebox[0.3cm][c]{0.019} & \cellcolor[HTML]{F9716D}\makebox[0.6cm][c]{1.112} & \makebox[0.3cm][c]{0.005} & \cellcolor[HTML]{FCB079}\makebox[0.6cm][c]{4.207} & \makebox[0.3cm][c]{0.050}  \\ \cline{1-20}
                              & MG                        & \cellcolor[HTML]{63BE7B}\makebox[0.6cm][c]{\textbf{56.09}} & \makebox[0.3cm][c]{3.45} & \cellcolor[HTML]{E2E383}\makebox[0.6cm][c]{57.05} & \makebox[0.3cm][c]{1.06} & \cellcolor[HTML]{FBB279}\makebox[0.6cm][c]{41.91} & \makebox[0.3cm][c]{4.27}  & \cellcolor[HTML]{63BE7B}\makebox[0.6cm][c]{\textbf{73.72}} & \makebox[0.3cm][c]{0.98} & \cellcolor[HTML]{7BC57D}\makebox[0.6cm][c]{61.62} & \makebox[0.3cm][c]{0.06} & \cellcolor[HTML]{FBAB77}\makebox[0.6cm][c]{49.30}  & \makebox[0.3cm][c]{0.01} & \cellcolor[HTML]{FEEA83}\makebox[0.6cm][c]{2.138} & \makebox[0.3cm][c]{0.100}   & \cellcolor[HTML]{F3E783}\makebox[0.6cm][c]{1.040}  & \makebox[0.3cm][c]{0.024} & \cellcolor[HTML]{FB9173}\makebox[0.6cm][c]{4.214} & \makebox[0.3cm][c]{0.009} \\
                              & SMILES                    & \cellcolor[HTML]{BAD780}\makebox[0.6cm][c]{54.98} & \makebox[0.3cm][c]{2.55} & \cellcolor[HTML]{EAE583}\makebox[0.6cm][c]{56.78} & \makebox[0.3cm][c]{0.11} & \cellcolor[HTML]{FBEA84}\makebox[0.6cm][c]{45.56} & \makebox[0.3cm][c]{12.63} & \cellcolor[HTML]{FEE783}\makebox[0.6cm][c]{49.26} & \makebox[0.3cm][c]{0.59} & \cellcolor[HTML]{C9DC81}\makebox[0.6cm][c]{55.88} & \makebox[0.3cm][c]{0.02} & \cellcolor[HTML]{63BE7B}\makebox[0.6cm][c]{\textbf{51.89}} & \makebox[0.3cm][c]{0.08} & \cellcolor[HTML]{DAE081}\makebox[0.6cm][c]{2.066} & \makebox[0.3cm][c]{0.071} & \cellcolor[HTML]{D3DE81}\makebox[0.6cm][c]{1.023} & \makebox[0.3cm][c]{0.006} & \cellcolor[HTML]{D3DE81}\makebox[0.6cm][c]{4.160}  & \makebox[0.3cm][c]{0.004} \\
                              & CCO              & \cellcolor[HTML]{B8D780}\makebox[0.6cm][c]{55.00}    & \makebox[0.3cm][c]{4.01} & \cellcolor[HTML]{63BE7B}\makebox[0.6cm][c]{\textbf{61.19}} & \makebox[0.3cm][c]{0.43} & \cellcolor[HTML]{93CC7E}\makebox[0.6cm][c]{54.11} & \makebox[0.3cm][c]{3.28}  & \cellcolor[HTML]{69C07C}\makebox[0.6cm][c]{72.94} & \makebox[0.3cm][c]{3.57} & \cellcolor[HTML]{63BE7B}\makebox[0.6cm][c]{\textbf{63.37}} & \makebox[0.3cm][c]{0.03} & \cellcolor[HTML]{E2E383}\makebox[0.6cm][c]{50.23} & \makebox[0.3cm][c]{0.08} & \cellcolor[HTML]{8CC97D}\makebox[0.6cm][c]{1.907} & \makebox[0.3cm][c]{0.064} & \cellcolor[HTML]{63BE7B}\makebox[0.6cm][c]{\textbf{0.963}} & \makebox[0.3cm][c]{0.004} & \cellcolor[HTML]{63BE7B}\makebox[0.6cm][c]{\textbf{4.074}} & \makebox[0.3cm][c]{0.006} \\
\multirow{-4}{*}{Ours} & EWA                & \cellcolor[HTML]{7AC57D}\makebox[0.6cm][c]{55.80}  & \makebox[0.3cm][c]{3.14} & \cellcolor[HTML]{75C37C}\makebox[0.6cm][c]{60.63} & \makebox[0.3cm][c]{0.55} & \cellcolor[HTML]{9ECF7F}\makebox[0.6cm][c]{53.17} & \makebox[0.3cm][c]{10.58} & \cellcolor[HTML]{CDDD82}\makebox[0.6cm][c]{57.44} & \makebox[0.3cm][c]{0.96} & \cellcolor[HTML]{73C37C}\makebox[0.6cm][c]{62.22} & \makebox[0.3cm][c]{0.09} & \cellcolor[HTML]{D5DF82}\makebox[0.6cm][c]{50.41} & \makebox[0.3cm][c]{0.01} & \cellcolor[HTML]{63BE7B}\makebox[0.6cm][c]{\textbf{1.823}} & \makebox[0.3cm][c]{0.060}  & \cellcolor[HTML]{81C67C}\makebox[0.6cm][c]{0.979} & \makebox[0.3cm][c]{0.019} & \cellcolor[HTML]{8FCA7D}\makebox[0.6cm][c]{4.108} & \makebox[0.3cm][c]{0.003} \\ \hline
\end{tabular}
\end{center}
\end{table*}

\subsubsection{Datasets}

\paragraph{Training Dataset}

We choose ZINC \cite{irwin2012zinc} as the training dataset, which contains 249,455 compounds. These compounds cover a wide range of chemical spaces, including drug samples, natural products, fragment compounds, prodrug molecules, and various ligands. We randomly split the training set and validation set in a 9:1 ratio. It is worth noting that the pre-training datasets for the selected backbones (CHEM-BERT and Grover-large) include ZINC, ZINC15 \cite{sterling2015zinc}, and Chembl \cite{gaulton2012chembl}. This ensures that the training data does not introduce new data outside the backbones, proving that the improvement of the fusion method is due to the complementarity of the two modalities rather than additional training data.

\paragraph{Downstream Task Datasets}
We select 6 classification tasks and 3 regression tasks from MoleculeNet. Classification tasks include BBBP, BACE, SIDER, Clintox, Toxcast, and Tox21. Regression tasks include ESOL, Lipo, and Freesolv. For both classification and regression tasks, we apply scaffold splitting \cite{hu2016computational}. The datasets are divided into training, validation, and test sets with a ratio of 80\%, 10\%, and 10\%, respectively.

\subsubsection{Baselines}


In this section, we select baselines from the perspective of model fusion methods. We choose three representative fusion methods to compare with our proposed method: 1. Simple Fusion: Aggregation of unimodal encodings without training. 2. Contrastive Learning Fusion: A multimodal molecular-level fusion method. 3. DMP Fusion \cite{zhu2021dual}: A fusion method that combines multimodal molecular-level learning with unimodal atomic-level learning.
Since aggregation operations are independent of the encoder training method, they can be combined with contrastive learning fusion, DMP fusion, and our method for comparison.

\paragraph{Simple Fusion}
\label{simple_fusion}

Simple fusion uses aggregation operations directly, without training, to integrate multimodal information. There are two proposed aggregation operations: (a) Element-Wise Addition (EWA) is an aggregation operation that adds each corresponding element of two embeddings. (b) Concatenation Operation (CCO) concatenates two embeddings along the specified dimension to form a single, longer embedding.

\paragraph{Contrastive Learning}

Comparative learning \cite{radford2021learning} is a common method for molecular modal fusion. As a baseline, we use contrastive learning to align different molecular modalities. Modalities of the same molecule are treated as positive samples, while those of different molecules are treated as negative samples.

\paragraph{DMP}
DMP \cite{zhu2021dual} is a typical work that integrates SMILES and molecule graphs, using atomic-level masking methods and dual-view molecular-level contrastive learning on two single modalities, respectively. Unfortunately, this approach trains two encoders from scratch, ignoring existing powerful single-modal encoders. As a result, it requires a large dataset of 110M training samples, about 4.4 times the size of our dataset, leading to unnecessary computational waste. Both DMP and our approach have a dual-tower structure. We replace the backbones in DMP with those selected for our method while retaining the DMP training approach.

\subsubsection{Metrics}

We use ROC-AUC to evaluate classification tasks and RMSE to evaluate regression tasks. In cases where datasets contain multiple tasks, we calculate the average ROC-AUC score. Additionally, we employ an early-stop mechanism based on the validation set loss and record the average performance of 3 times.

\subsection{Results}

We validate our proposed fusion method on drug property prediction tasks, and the experimental results are shown in Table \ref{tab:performance_comparison}. Our fusion method achieves significant performance improvements.

Unlike previous work \cite{zhu2021dual, hou2022graphmae, zhu2022featurizations}, our method fully utilizes two encoders of different modalities. Previously, in downstream tasks, only a single encoder was used, which is insufficient for utilizing molecular multimodal complementary information: using only one encoder means that there is only one modality to supplement the information of another modality without the reverse information. Therefore, our work investigates the complementary effect of information between two molecular modalities. We evaluate the SMILES encoder only, molecule graph encoder only, EWA, and CCO under the conditions of no-train, contrastive learning, DMP, and our proposed method, as shown in Table \ref{tab:performance_comparison}.

\begin{figure*}[!htbp]
    \centering
    \includegraphics[width=\textwidth]{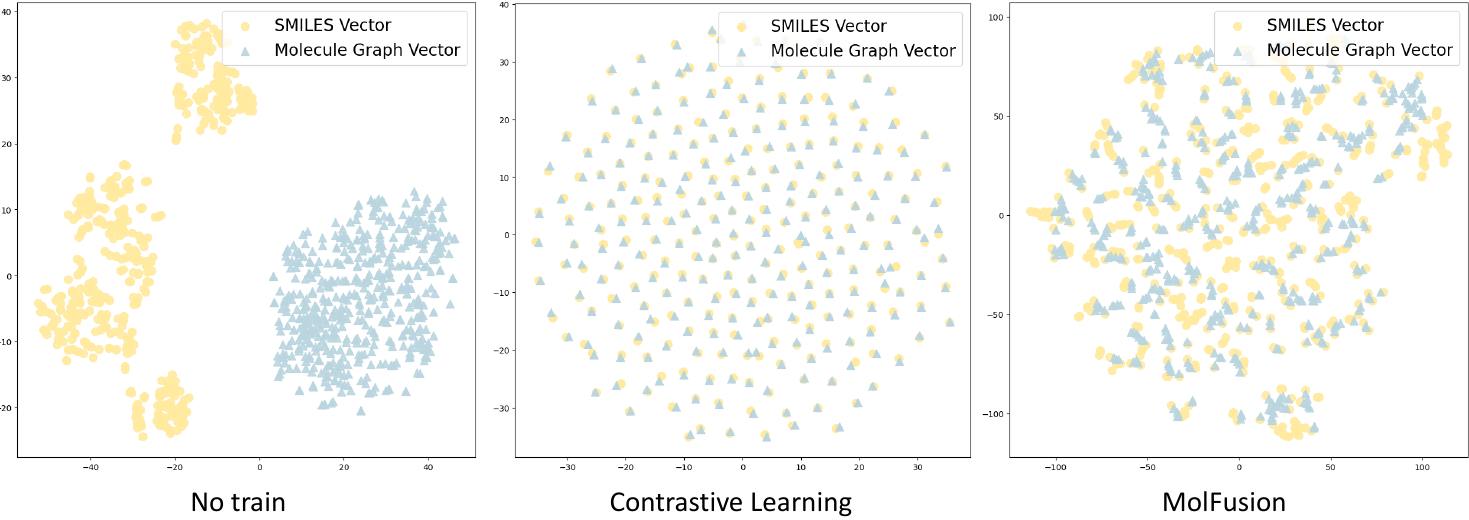}
    \caption{Visualization of molecular vectors from the ZINC dataset under three conditions: (1) No train: The vector spaces of different molecular representations do not overlap without training. (2) Contrastive Learning: The vector spaces between different representations are highly overlapping, leading to the loss of complementary information between the two modalities. (3) MolFusion: Our method results in partially overlapping vector spaces, consistent with the assumption that the information of different representations of the same molecule is partially duplicated, and the unique information of each modality is complementary. The SMILES vectors are represented in yellow, and the molecule graph vectors are represented in blue.}
\label{vis}
\end{figure*}

The results in Table \ref{tab:performance_comparison} show that, among all the compared fusion methods, our approach achieves the most significant improvement by aggregating multimodal representations, as opposed to relying solely on single-modal representation. For example, in the BBBP dataset, compared to the maximum performance of the encoder-only method, our method shows an improvement of 4.64\% using aggregation operations, while the no-train condition results in a decrease of 1.60\%, contrastive learning achieves an improvement of 2.93\%, and DMP shows a minor increase of 0.12\%. This clearly demonstrates the effectiveness of our method in leveraging complementary multimodal information. Compared to our method, the effectiveness of other fusion methods decreases to varying degrees. Comparing DMP with our method, we can analyze that multimodal atomic-level methods can better integrate molecular multimodal information than single-mode atomic-level methods.

\subsection{Ablation}

In order to better understand the roles of the two components in MolFusion, we conduct ablation experiments. We run MolSim and AtomAlign components separately and combine them with atomic-level masked learning model and molecular-level contrastive learning, respectively, to form a multi-granularity method. We select SIDER, BBBP, Tox21, and Toxcast as the ablation experimental datasets and record the maximum experimental result of four aggregation operations as the result of each fusion method. The results demonstrate the effectiveness of two components in our method.

\begin{figure}[h]
    \centering
    \includegraphics[width=0.48\textwidth]{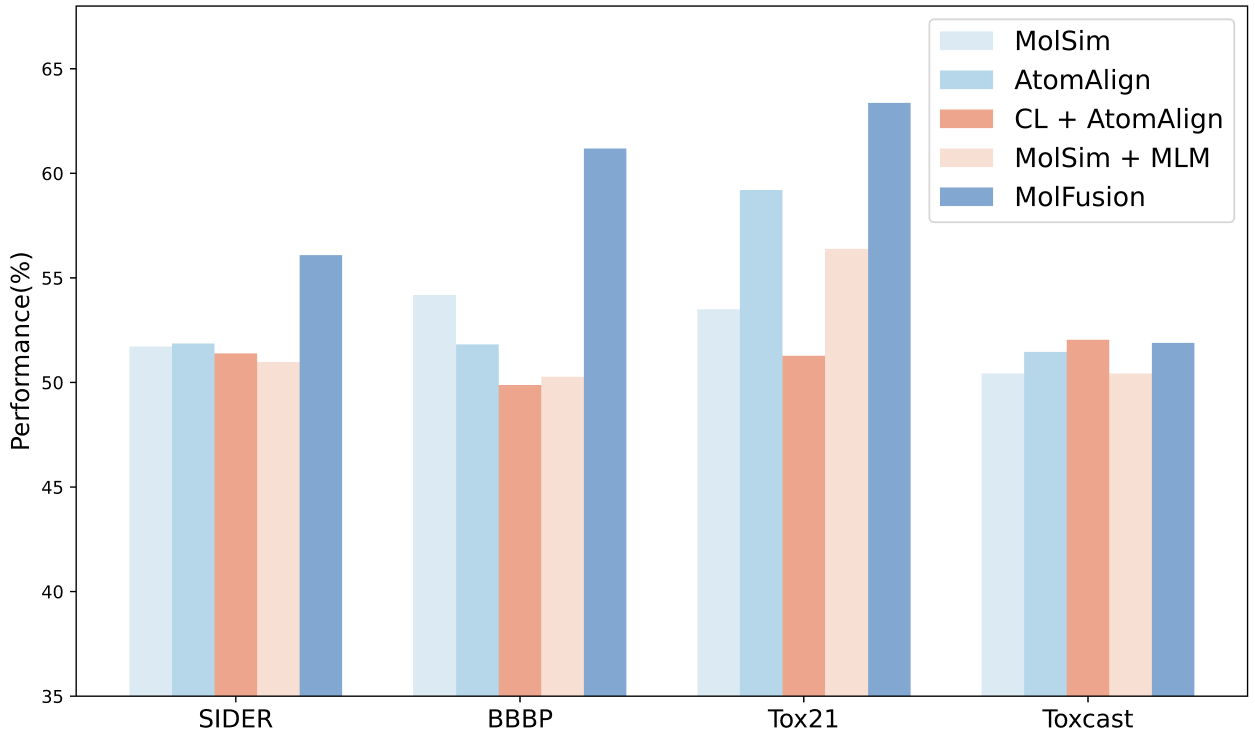}
    \caption{The performance of various ablation methods across different datasets.}
\label{ablation}
\end{figure}

As shown in Figure \ref{ablation}, the experimental results show that our method is optimal among many combinations of molecular-level and atomic-level methods. Neither MolSim nor AtomAlign components alone can effectively enable the model to learn the complementary information from molecular modalities. For instance, in the SIDER dataset, MolSim alone achieves a ROC-AUC of 54.23\% and AtomAlign 50.21\%, but combined in our method, the ROC-AUC improves to 56.09\%. In addition, whether we replace the molecular-level component in our method with contrastive learning or replace the atomic-level component in our method with a single-modal masked learning model, the effect decreases to varying degrees. For example, in the Tox21 dataset, replacing the molecular-level and atomic-level components in our approach with single-modal methods results in a performance drop of 12.09\% and 6.98\%, respectively. These prove the rationality and effectiveness of the two components in MolFusion, confirming that multi-granularity fusion method specifically designed for multimodal data can more effectively learn complementary information between different modalities.

\subsection{Visualization of Fusion Effects}
\label{sec:vis_effects}

To visually demonstrate the effectiveness of our proposed fusion method in integrating two molecular representations, we use t-SNE for dimensionality reduction visualization. We select 500 molecular vectors from the ZINC dataset and visualize them under three conditions: no-train, contrastive learning, and our method. The results validate the \textbf{Complementary Information Assumption}.


As shown in Figure \ref{vis}, our method results in partially overlapping vector spaces, consistent with \textbf{Complementary Information Assumption}. This leads to the most significant improvement when fusing representations of the two modalities.

%% file: sections/5-conclusion.tex
\section{Conclusion}
In conclusion, we propose MolFusion, a novel multimodal and multi-granularity fusion method. This method can better learn complementary information between different modalities through the molecular-level method MolSim and the atomic-level method AtomAlign. We achieve significant performance improvements on multiple classification and regression tasks in MoleculeNet. The effectiveness of our method in integrating different molecular modalities is verified through the ablation experiment, and the rationality of our method is further demonstrated through dimensionality reduction visualization.